\documentclass{wscpaperproc}
\usepackage{latexsym}
\usepackage[normalem]{ulem}
\usepackage{graphicx}
\usepackage{mathptmx}
\usepackage[T1]{fontenc}
\usepackage{tikz}
\usetikzlibrary{arrows.meta, positioning, shadows, fit, backgrounds, calc}   

\usepackage{tikz-cd}
\usepackage{multicol}
\usepackage{multirow}
\usepackage{array}
\usepackage{tabularx}

\usepackage{amsmath}
\usepackage{xcolor}

\definecolor{primary}{RGB}{26, 95, 122}         
\definecolor{primarylight}{RGB}{240, 247, 248}  
\definecolor{accent}{RGB}{226, 62, 87}          
\definecolor{accentbg}{RGB}{255, 246, 246}      
\definecolor{textdark}{RGB}{34, 40, 49}         

\definecolor{operational}{HTML}{9B59B6}
\definecolor{analysis}{HTML}{3498DB}
\definecolor{coordination}{HTML}{E67E22}
\definecolor{verifier}{HTML}{2ECC71}
\definecolor{response}{HTML}{E74C3C}

\usepackage{amsmath}
\usepackage{amsfonts}
\usepackage{amssymb}
\usepackage{amsbsy}
\usepackage{amsthm}

\usepackage{booktabs}   

\newcommand{\nomark}{--}

\usepackage[pdftex,colorlinks=true,urlcolor=blue,citecolor=black,anchorcolor=black,linkcolor=black]{hyperref}

\newtheoremstyle{wsc}
{3pt}
{3pt}
{}
{}
{\bf}
{}
{.5em}
{}

\theoremstyle{wsc}

\begin{document}

%
%

\pagestyle{fancyplain}

\thispagestyle{plain}
\firstPageHead{}

\chead{\fancyplain{}{\itshape Flandre, Nwala, and Giabbanelli}}

\rhead{}
\cfoot{}
\renewcommand{\headrulewidth}{0pt} 

\makeatletter
\let\@internalcite\cite
\def\cite{\def\@citeseppen{-1000}%
    \def\@cite##1##2{(##1\if@tempswa , ##2\fi)}%
    \def\citeauthoryear##1##2##3{##1 ##3}\@internalcite}
\def\citeNP{\def\@citeseppen{-1000}%
    \def\@cite##1##2{##1\if@tempswa , ##2\fi}%
    \def\citeauthoryear##1##2##3{##1 ##3}\@internalcite}
\def\citeN{\def\@citeseppen{-1000}%
    \def\@cite##1##2{##1\if@tempswa, ##2)\else{}\fi}%
    \def\citeauthoryear##1##2##3{##1 (##3)}\@citedata}
\def\citeA{\def\@citeseppen{-1000}%
    \def\@cite##1##2{(##1\if@tempswa , ##2\fi)}%
    \def\citeauthoryear##1##2##3{##1}\@internalcite}
\def\citeANP{\def\@citeseppen{-1000}%
    \def\@cite##1##2{##1\if@tempswa , ##2\fi}%
    \def\citeauthoryear##1##2##3{##1}\@internalcite}
\def\shortcite{\def\@citeseppen{-1000}%
    \def\@cite##1##2{(##1\if@tempswa , ##2\fi)}%
    \def\citeauthoryear##1##2##3{##2 ##3}\@internalcite}
\def\shortciteNP{\def\@citeseppen{-1000}%
    \def\@cite##1##2{##1\if@tempswa , ##2\fi}%
    \def\citeauthoryear##1##2##3{##2 ##3}\@internalcite}
\def\shortciteN{\def\@citeseppen{-1000}%
    \def\@cite##1##2{##1\if@tempswa, ##2\else{}\fi}%
    \def\citeauthoryear##1##2##3{##2 (##3)}\@citedata}
\def\shortciteA{\def\@citeseppen{-1000}%
    \def\@cite##1##2{(##1\if@tempswa , ##2\fi)}%
    \def\citeauthoryear##1##2##3{##2}\@internalcite}
\def\shortciteANP{\def\@citeseppen{-1000}%
    \def\@cite##1##2{##1\if@tempswa , ##2\fi}%
    \def\citeauthoryear##1##2##3{##2}\@internalcite}
\def\citeyear{\def\@citeseppen{-1000}%
    \def\@cite##1##2{(##1\if@tempswa , ##2\fi)}%
    \def\citeauthoryear##1##2##3{##3}\@citedata}
\def\citeyearNP{\def\@citeseppen{-1000}%
    \def\@cite##1##2{##1\if@tempswa , ##2\fi}%
    \def\citeauthoryear##1##2##3{##3}\@citedata}
%
%
%
\def\@citedata{%
    \@ifnextchar [{\@tempswatrue\@citedatax}%
                  {\@tempswafalse\@citedatax[]}%
}

\def\@citedatax[#1]#2{%
\if@filesw\immediate\write\@auxout{\string\citation{#2}}\fi%
  \def\@citea{}\@cite{\@for\@citeb:=#2\do%
    {\@citea\def\@citea{, }\@ifundefined
       {b@\@citeb}{{\bf ?}%
       \@warning{Citation `\@citeb' on page \thepage \space undefined}}%
{\csname b@\@citeb\endcsname}}}{#1}}%

%
\def\@citex[#1]#2{%
\if@filesw\immediate\write\@auxout{\string\citation{#2}}\fi%
  \def\@citea{}\@cite{\@for\@citeb:=#2\do%
    {\@citea\def\@citea{; }\@ifundefined
       {b@\@citeb}{{\bf ?}%
       \@warning{Citation `\@citeb' on page \thepage \space undefined}}%
{\csname b@\@citeb\endcsname}}}{#1}}%

%
\def\@biblabel#1{}
\makeatother



\newdimen\bibindent
\bibindent=0.0em
\def\thebibliography#1{\section*{\refname}\list
   {}{\settowidth\labelwidth{[#1]}
   \leftmargin\parindent
   \itemindent -\parindent
   \listparindent \itemindent
   \itemsep 0pt
   \parsep 0pt}
   \def\newblock{}
   \sloppy
   \sfcode`\.=1000\relax}


\setlength{\baselineskip}{12.7pt}

\title{Composing Verifiable Conceptual Models via Building Blocks:\\Towards Design-Time Verification of Agentic AI Workflows}

\author{\begin{center}Noé Y. Flandre\textsuperscript{1}, Alexander C. Nwala\textsuperscript{2}, and Philippe J. Giabbanelli\textsuperscript{3}\\
[11pt]
\textsuperscript{1}Team EVERGREEN, Inria, Centre Inria d'Universit\'e C\^ote d'Azur, Montpellier, France\\
\textsuperscript{2}Department of Data Science, William \& Mary, Williamsburg, VA, USA\\
\textsuperscript{3}Office of Enterprise Research and Innovation, Old Dominion University, Norfolk, VA, USA\\
\end{center}
}

\maketitle

\vspace{-12pt}

\section*{ABSTRACT}
Agentic AI systems orchestrate multiple LLM-based agents through workflow architectures that coordinate decisions, tools, and external actions. While current platforms emphasize runtime safeguards, little support exists for verifying workflows during system design. From a Modeling \& Simulation perspective, this gap is analogous to composing conceptual models without verifying whether their building blocks interact coherently. We propose a design-time verification approach that models agentic workflows as compositions of reusable building blocks and checks their compatibility through twelve structural rules. We implemented these rules in a software prototype and evaluated them using two openly released datasets: 48 workflows with known design flaws and 168 variants that preserve workflow logic but alter graph structure. Results show that our verifier reliably detects violations even when flawed designs are obscured through structural transformations such as splitting tasks between agents. Future works could combine our verification with community repositories of building blocks to compose safe agentic workflows.

\section{INTRODUCTION}
\label{sec:intro}

Agentic AI systems consist of collections of autonomous agents powered by large language models (LLMs) that plan, act, and interact with limited human oversight. These systems increasingly support customer-facing and operational tasks (e.g., chatbots, automated decision pipelines, tool-using assistants), and they are being extended to settings that involve complex coordination and third-party agents. In such contexts, agentic AI systems face substantial cybersecurity and safety risks. Some risks come from malicious or inauthentic actors, particularly when executing tools over untrusted data~\shortcite{debenedetti2024agentdojo}. Risks for leakage, misuse (e.g., fraud, cybercrime, harassment), and unsafe delegation have been reported in recent studies~\shortcite{zhan2024injecagent}. Although studies often cover adversarial settings, risks exist even without attacks: agents can (inadvertently) leak proprietary data through APIs or have permissions and privileges that exceed their intended roles. In their empirical analysis of 1642 execution traces, \shortciteN{pan2025why} found that 44.2\% of issues were due to poor system design (e.g., incorrect or ambiguous specifications of prompts, roles, or workflows), 32.3\% came from poor alignment between agents, and 23.5\% were verification-related errors (i.e., verification was missing, incomplete, or incorrect). In this paper, we address these challenges from a Modeling \& Simulation (M\&S) perspective. Prompt engineering has been extensively studied in prior works, such as injection and jailbreak attacks \shortcite{knowlton2026prompt,duarte2026systematic}, or best practices in the M\&S context~\cite{giabbanelli2026guide}. Rather than prompts, we focus on \textit{how the composition of agents, their interactions, and their control flows can introduce design issues in agentic workflows}. 

Our proposed approach seeks to make agent behaviors explicit and verifiable so that issues are detected and fixed at the design stage, thus avoiding unintended or irreversible outcomes during deployment. Our work thus covers the three themes from \shortciteN{pan2025why}: we contribute to system design from a workflow perspective, introduce a structural coordination layer to support alignment between agents, and articulate a set of structural rules to enable design-time verification. This approach is needed because \textit{current agentic AI ecosystems provide limited support for identifying unsafe compositions during system design}. Widely used frameworks such as \href{https://github.com/anthropics/claude-agent-sdk-python}{Claude Agent SDK}, \href{https://github.com/openai/openai-agents-python}{OpenAI Agents SDK}, \href{https://crewai.com/}{CrewAI}, \href{https://www.microsoft.com/en-us/research/project/autogen/}{Microsoft AutoGen}, and \href{https://aws.amazon.com/bedrock/agentcore}{AWS Bedrock AgentCore} primarily rely on runtime safeguards (e.g., guardrails, policy enforcement, or self-evaluation) at deployment time, which prevent cascading or irreversible behaviors once execution has begun~\cite{borghoff2025beyond}. Other tools, including orchestration frameworks such as \href{www.vellum.ai}{Vellum}, \href{https://www.langchain.com/langgraph}{LangGraph}, DSPy~\shortcite{khattab2023dspy}, and \href{https://cloud.google.com/products/agent-builder}{Google Vertex AI Agent Builder}, provide partial interface- or tool-level validation but do not reason about the semantic compatibility of agent behaviors or the correctness of multi-agent control flows. Visual orchestration environments such as Rivet~\cite{ironclad2024rivet} improve debugging and iteration, yet they still lack automated mechanisms to verify that agent compositions are coherent, safe, or meaningful prior to execution. These limitations echo research observations that unsafe chains can be created without any design-time warning~\shortcite{akshathala2025beyond,bommasani2022opportunitiesrisksfoundationmodels}. Consequently, our work contributes to addressing the research gap of fine-grained verification approaches to identify incompatible or unsafe interactions before investing in building problematic workflows (detailed in Section~\ref{sec:background}), and certainly before executing them in ways that may trigger cascading effects.

Two strands of M\&S research guide our work. First, empirical analyses have shown that many Agent-Based Models are repeatedly implemented from scratch even though they rely on recurring components. This indicates that mechanisms can be formalized as \textit{reusable building blocks} \cite{cheng2023identifying}. These building blocks are typically defined as self-contained, interoperable components that can be reused, provided that their integration satisfies requirements such as semantic consistency and clear interfaces \shortcite{berger2024towards}. Prior works proposed using these blocks to design ABMs (e.g., a block for initialization, another for agent movements), which would speed-up development and improve code quality instead of potentially re-implementing a common component with bugs~\shortcite{schroeder2022towards}. In this paper, we take this approach to specify Agentic AI workflows, not from a concern of development speed (e.g., current frameworks and vibe coding practices allow to quickly create workflows) but with a focus on verification. Second, formal analyses of simulation model specifications show that many verification problems are computationally difficult or even undecidable. For example, determining whether a model implementation satisfies a specification or whether certain states occur during execution can be NP-hard, while other properties such as model equivalence may be undecidable \cite{page1999observations}. Previous works thus verified the composition of elements in agent-based simulations through \textit{sets of rules}, for instance by ensuring that an agent could not be asked to perform incompatible requests in the same cycle such as changing its value by a relative and into an absolute amount~\cite{giabbanelli2019cofluences}. Our work expands on this approach by formalizing relationships between component to identify invalid combinations of mechanisms, while also suggesting missing mechanisms (e.g., safeguard agents).

Our main contribution is to design, implement, and evaluate a rule-based verification framework for agentic AI workflows that models agents and coordination mechanisms as reusable building blocks and detects unsafe compositions at design time. As detailed in Section~\ref{sec:methods}, our approach (i) formalizes agentic workflows as compositions of interoperable mechanisms, and (ii) introduces structural and semantic rules to verify the compatibility of these mechanisms. Then, our implementation and evaluation in Section~\ref{sec:results} (iii) demonstrates how such verification can detect invalid or unsafe workflow configurations prior to execution. The implications of these results for building resilient agentic ecosystems are examined in Section~\ref{sec:discussion} along with priorities for future works. 

\section{BACKGROUND}
\label{sec:background}

\subsection{Principles of Agentic AI Workflows}
\label{sec:principles}

As there is no standard or regulation when it comes to agentic AI, the term may be used loosely to cover vastly different architectures, which call for different approaches to verification. For example, we may occasionally encounter solutions labeled as `agentic' despite consisting of a series of prompts, performed one after the other as would have been the case several years ago. In such cases, there is only one instance of an agent with each role: for example, one agent extracts a conceptual model from text, the next agent turns it into code, and so on. In this paper, we focus on modern agentic AI systems that can \textit{involve more complex delegation mechanisms such as branching patterns, when an agent calls several others} (e.g., a modeler-centric agent notifies that simulation runs are done and logged into a database, and at the same time a stakeholder-facing agent provides a summary of the results in context). 

\begin{figure}[h]
\centering
\includegraphics[width=1\textwidth]{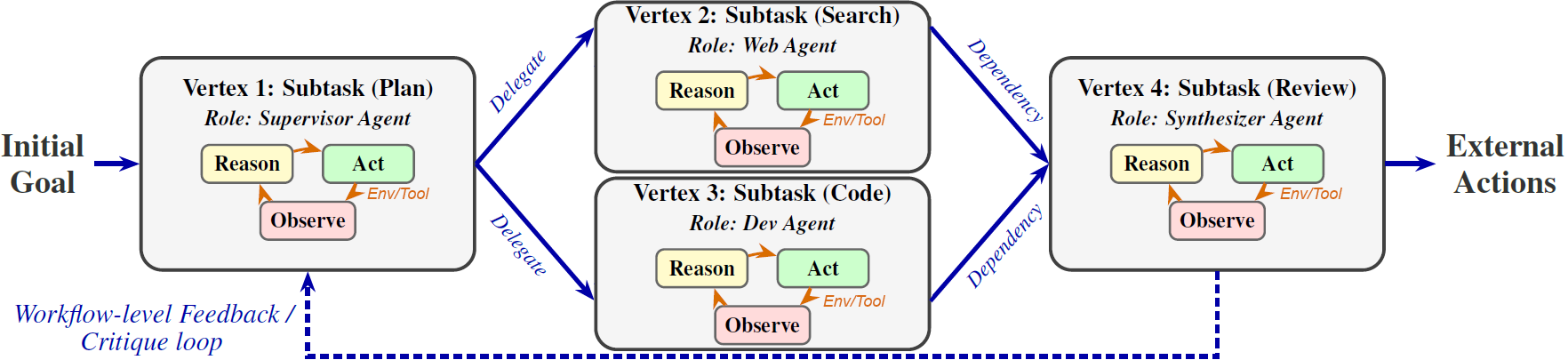}
\caption{Example of Agentic AI workflow architecture. Vertices represent subtasks assigned to role-specialized agents. Each agent executes an internal \emph{reason--act--observe} micro-loop (orange arrows) to interact with tools and the environment. Workflow dependencies dictate the macro-level pipeline (solid blue arrows), alongside an optional workflow-level feedback loop (dashed blue arrow) for iterative critique.}
\label{fig:architecture}
\end{figure}

Agentic AI workflows specify how one or more LLM-based agents transform an initial goal into a sequence of decisions and external actions \cite{khamis2025agentic}. For example, \shortciteN{niu2025flow} represent an agentic AI workflow as a directed acyclic graph (named `activity-on-vertex'), where vertices are subtasks, directed edges encode dependency constraints between subtasks and lastly a set of role-specialized agents assigned to execute subsets of these subtasks. Many practical agentic frameworks represent workflows as (stateful) directed graphs with explicit feedback cycles (i.e., not necessarily acyclic) to support iterative replanning and tool-use loops \cite{langchain2024langgraph}. These representations encode branching, parallelism and termination conditions, thus going beyond a linear prompt chain. Within a vertex, an agent can perform multiple actions to satisfy its role. For example, agents can use the \textit{(reason, act, observe)} approach repeatedly, which combines internal deliberation with tool calls as well as environment interactions to decide when to end the loop and revise their own local plan to determine subsequent actions \cite{yao:react23}. Most workflows generally adopt role-specialized agents which coordinate through explicit interaction patterns (e.g delegation, critique, or supervisor worker loops), thus replicating a structured negotiation over shared artifacts. Figure \ref{fig:architecture} provides an overview of an Agentic AI workflow.

\subsection{Verifying Agentic AI Workflows}
Low code platforms for agentic AI workflows enable rapid prototyping without providing mechanisms to verify structural correctness, behavioral safety or semantic compatibility prior to execution. For instance, Flowise, a low-code drag-and-drop LLM workflow builder, does not provide guardrails or verification mechanisms. Policy checks are enforced via external gateways (e.g., TrueFoundry
AI Gateway). Similarly, Dify does not directly verify a workflow: it relies on external plugins like Azure AI Content Safety for moderation \cite{dify2025azure}. Among the tools that do provide verifications, capacities vary noticeably. Across 10 professional platforms for agentic AI, \textit{all provide verification at runtime, except AutoGen} which primarily delegates runtime safety to human-in-the-loop interventions (e.g., needing explicit user approval before executing generated code) or sandboxed execution environments \cite{microsoft2023autogen}. The main difference concerns verification \textit{at design time}. To enable comparisons, we consider three categories of design time verification: \textit{schema} (or `structural' verification) checks whether agent connections and input/output compatibility; \textit{behavioral} (or `semantic' verification) checks constraints on expected agent behavior (e.g safety policies, assertions); and \textit{multi-agent} assesses inter-agent message routing and coordination issues (e.g unreachable agents, deadlocks). As shown in Table \ref{tab:verification}, no platform currently covers all three aspects.

\begin{table*}[ht]
\centering
\caption{Pre-deployment verification mechanisms across industrial agentic AI platforms. \nomark~=~absent.}
\small
\begin{tabular*}{\textwidth}{@{\extracolsep{\fill}}l ccc@{}}
\toprule
\textbf{Platform}
  & \textbf{Schema}
  & \textbf{Behavioral}
  & \textbf{Multi-Agent} \\
\midrule
Claude SDK (Anthropic)       & \nomark   & \nomark   & \nomark   \\
Vellum.AI                    & Interface validation & Test suite & \nomark   \\
Rivet (Ironclad)             & Visual type checking   & Trivet testing & \nomark   \\
AWS Bedrock AgentCore        & OpenAPI blueprint check   & \nomark & \nomark   \\
LangGraph (LangChain)        & Pydantic types (compile-time) & \nomark   & Graph compilation   \\
CrewAI                       & Pydantic models   & \nomark   & CLI structure validation \\
OpenAI Agents SDK            & JSON Schema   & \nomark   & \nomark   \\
Google Vertex AI ADK         & \nomark & \nomark & \nomark   \\
DSPy (Stanford NLP)          & \nomark   & \nomark & \nomark   \\
AutoGen (Microsoft)          & \nomark   & \nomark   & \nomark   \\
\bottomrule
\end{tabular*}
\label{tab:verification}
\end{table*}

The prevailing industry approach to safety is the runtime guardrail, that is, auxiliary models that intercept inputs and outputs during execution to filter harmful content or block unauthorized actions \shortcite{rebedea:nemo23}. While necessary, runtime guardrails have limitations: they act only after a potentially unsafe instruction has been made \shortcite{wang2025sok_jailbreak_guardrails}; they may rely on smaller LLMs or classifiers to judge content which are subject to the same stochastic failures and adversarial attacks (e.g., `jailbreaking') as the agent they protect \shortcite{shang2025evolving_security_llms}; and  runtime filters can only check content, not flawed structure (e.g infinite loop, a `human in the loop' intervention being bypassed by a fallback logic) \shortcite{cemri2025whyfail}.

Research on verifying agentic AI workflows uses formal representations. For example, \citeN{meijer2026guardians} proposes that an agent first generate a structured execution plan describing tool calls and control flow (e.g., represented as a JSON abstract syntax tree), which is then statically \textit{verified before execution} using preconditions, postconditions, and invariants as well as data-flow analysis to prevent unsafe paths such as confidential data flowing from a retrieval tool to an external communication tool. For~\citeN{borghoff2025beyond}, the representation combines Colored Petri Nets (to encode coordination) with topic-annotated tokens for semantic processing \cite{borghoff2025beyond}. This representation enables formal reasoning about coordination properties such as valid task sequencing or deadlock-free execution within a workflow. By contrast, we focus on design-time verification of \textit{composability}. By operating at the \textit{mechanism composition level}, our goal is to detect issues before a workflow is fully written (thus saving time for modelers). For example, we may detect that two agents interact but no arbitration mechanism exists, which would trigger a rule ``a negotiation pattern requires a mediator or decision rule.''

\subsection{Application Domain: Agentic AI and Social Media}
\label{sec:casestudies}

Since the previous subsection covers approaches to verify agentic AI workflows, we need concrete workflows on which these tools can be tested. Online platforms provide a particularly relevant domain for such workflows. About 4.8 billion people globally use social media to consume and produce content, often without the editorial gatekeeping mechanisms typical of traditional media. While this openness encourages participation, it also allows harmful or illegal content (e.g., hate speech, graphic violence) to proliferate. Moderation strategies differ across platforms: for example, Meta explicitly prohibits cyberbullying and certain forms of harmful content on Facebook and Instagram~\cite{fb2025contentmod}, whereas other platforms adopt minimal moderation policies~\cite{parler2024contentmod}. Wikipedia offers another governance model in which automated bots address obvious violations such as vandalism while human editors enforce broader policies such as neutrality and consensus. Since millions of posts are created hourly~\cite{domo2024}, AI systems orchestrated as workflows automatically analyze posts, detect policy violations, and remove or escalate harmful content~\cite{meta2025aicontentmod}. These \textit{governance processes involve multiple agents, decision stages, and escalation mechanisms, making them suitable case studies for analyzing the structural correctness of agentic workflows}. In this section, we focus on three representative scenarios.

\begin{figure*}[t]
\centering
\includegraphics[width=1.0\textwidth]{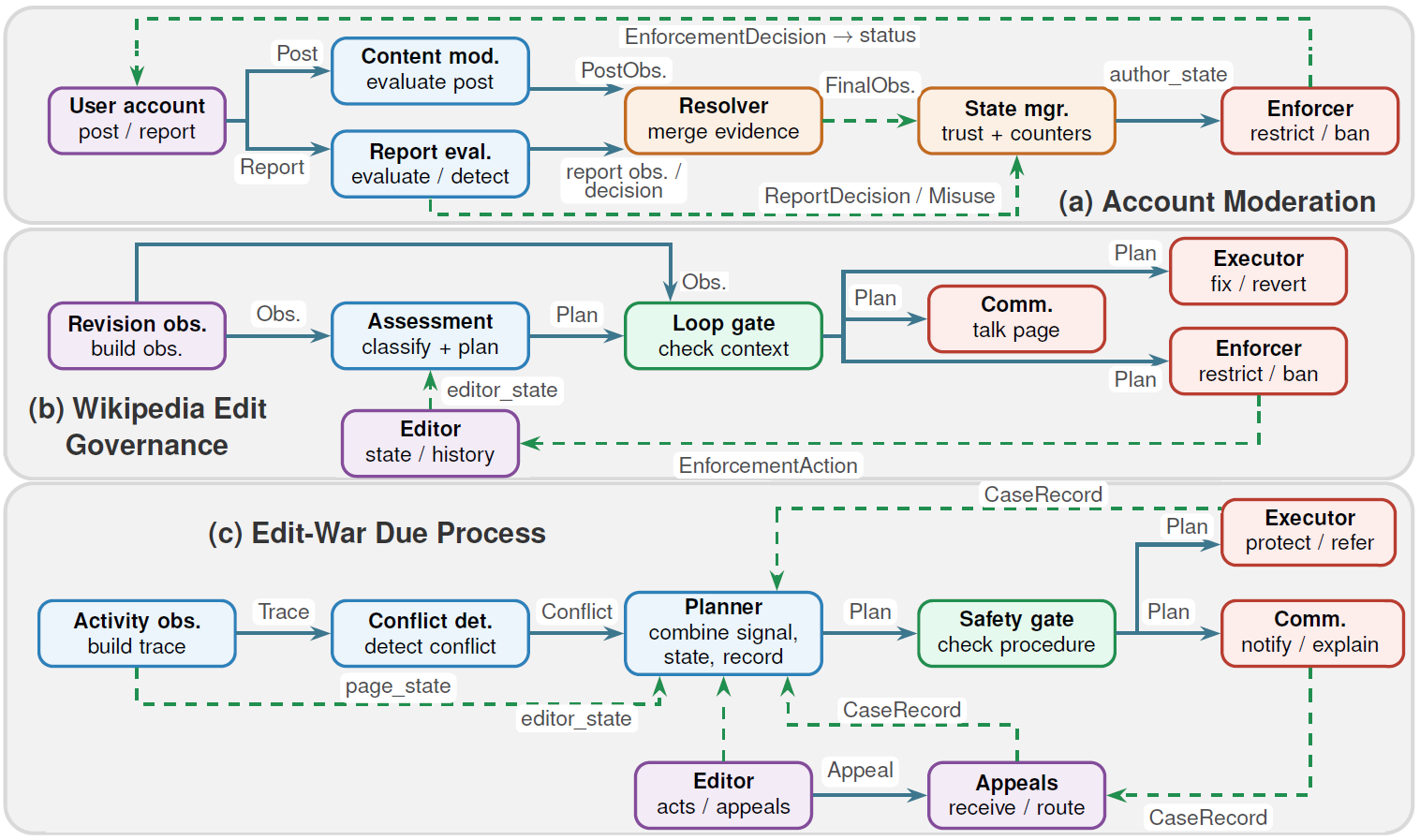}
\caption{Reference workflows for the three case studies. Solid blue arrows show
  control or plan flow; dashed green arrows show contextual continuity links,
  including state, feedback, and case-record propagation.}
\label{fig:workflows-overview}
\end{figure*}

The first case study (Figure~\ref{fig:workflows-overview}a) models account moderation and trust management on a social platform. Users generate posts (e.g., a tweet) and reports (e.g., they can report another user for violations), which are analyzed by a content moderator and a report evaluator, respectively. These moderators emit observations (e.g., minor or major violation). Because posts and reports can generate multiple streams of evidence, an evidence resolver aggregates them into a single assessment. A state manager then updates account-level variables such as trust score or violation counters, and a policy enforcer decides whether to restrict or ban the account. This workflow exemplifies how agents need to coordinate and each has a specialized role. Design issues may arise if trust updates are triggered directly from raw observations rather than aggregated evidence (updating trust from each signal \textit{independently} could penalize/reward an account several times for the same post) or if enforcement actions are triggered directly from observations (a decision stage should interpret moderation labels in the contest of prior violations to avoid inconsistent sanctions).

The second case study (Figure~\ref{fig:workflows-overview}b) focuses on Wikipedia edit governance and quality control. The focus is on the collaborative maintenance of knowledge in an open editing environment, where edits must be evaluated for quality, intent, and compliance with community norms. In this workflow, an editor (human or bot) edits a page, which is detected by an observer (e.g., a human who uses the built-in `watchlist' feature to be notified or external tools for real-time monitoring, or a bot). The quality of the edits is evaluated (clean, minor issue, serious problem such as vandalism) leading to a coordinated response including editing actions (e.g., reverting vandlism), communicating with the editor (through their `talk' page), or enforcement actions (e.g., temporary ban). Design issues include `bot wars', so an Agentic AI workflow should have a loop-safety mechanism to verify that proposed actions would not simply undo previous automated edit.

The third case study (Figure~\ref{fig:workflows-overview}c) models edit war detection and dispute resolution on Wikipedia, when editors repeatedly modify or revert each other’s contributions. Unlike the previous workflows, which focus primarily on content quality, this scenario addresses governance processes involving escalation, transparency, and due process. The workflow begins with an agent that tracks revision patterns and constructs interaction traces summarizing repeated reversions or other conflict indicators. An agent specialized in conflict detection then analyzes these traces and produces signals describing the severity of the conflict. A planning agent creates a structured plan combining possible responses (e.g., warnings, page protections, mediation requests) which are checked (i.e., they are \textit{gated}) by a safety agent, for example, ensuring that evidence exists for the conflict and that affected users retain the ability to appeal. Finally, an agent with elevated permission implements page or user restrictions, while an agent with minimal permissions informs participants. Potential design flaws include inconsistent conflict management or a lack of due process: if a user have write permissions to appeal, and banning prevents the account for writing on all pages, then the appeal is effectively unreachable.

\section{METHODS}
\label{sec:methods}

Our process is summarized in Figure \ref{fig:workflows-overview}. For each of the three case studies (Section~\ref{sec:casestudies}), we formalize agent interactions through a library of reusable behavioral building blocks, each encapsulating a distinct function (e.g, planning), and exposing typed inputs, outputs, and semantic flags that enable structural reasoning about control flow and data dependencies. To demonstrate the ability of our approach at identifying and resolving design flaws, we take a valid (or `reference') workflow for each case study and mutate it to obtain a total of 48 workflows. We verify each workflow through twelve cross-cutting generic structural rules, inspired by best practices in agent-based modeling (e.g., avoiding incompatible concurrent modifications). Since our dataset provides a `ground truth' of design errors, our evaluation builds a confusion matrix to record how many errors we correctly or incorrectly identified and fixed. From there, we derive and report classic measures (e.g., precision, recall, mean absolute error, root mean square error).

\begin{figure}[h]
\centering
\includegraphics[width=1\textwidth]{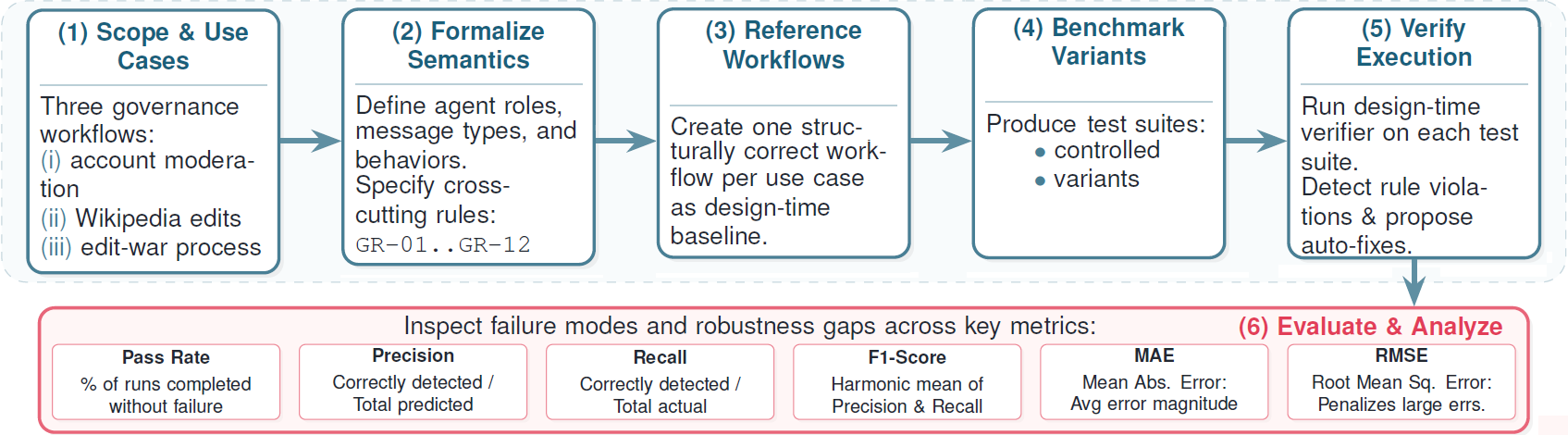}
\caption{Starting from three complementary agentic AI use cases, we formalize workflow semantics and cross-cutting rules, construct a ground truth dataset of workflows and variants with labeled errors, run verification on each variant, and evaluate performance through detection and error-calibration metrics.}
\label{fig:method-overview}
\end{figure}

\subsection{An ABM-inspired representation and set of rules}
\label{sec:representation}

Section~\ref{sec:principles} described how modern agentic AI workflows are typically represented as graphs of interacting agents and subtasks. While such representations capture control flow (e.g., branching, delegation, or feedback loops), they often treat the internal semantics of workflow components as opaque. As a result, the meaning of interactions, state changes, and decision logic remains implicit, which complicates verification. To address this limitation, we represent agentic workflows as systems of \textit{stateful agents performing semantically typed behaviors}. This perspective is inspired by ABMs, where interacting entities have persistent state, explicit interactions, and rule-governed behavior. Formally, we represent an agentic workflow as $W = (A, S, M, B, E, G)$ where $A$ is a set of agent roles, $S$ a set of state variables, $M$ a set of message types, $B$ a set of behavioral blocks, $E$ a set of execution edges, and $G$ a set of guards. 

The \textbf{agents} $A$ aren't limited to temporary prompt executors: each agent role $a \in a$ maintains a subset of the state variables $S$ as part of its persistent internal state (e.g., trust scores, violation counters, or page status). Making the \textbf{state variables} explicit at the workflow level allows verification rules to reason about which behaviors may read or modify them. Agents interact through \textbf{typed messages} $M$ (e.g., observations, reports, plans, enforcement decisions). Typing messages makes the meaning of interactions explicit: an observation produced by an observer agent is distinct from a plan produced by a planner or an enforcement decision produced by a policy agent. This structure parallels event-driven simulation frameworks in which agents communicate through explicitly defined event types. Because message types encode the semantics of information flows, they enable verification rules that prevent incompatible interactions.

The \textbf{behavioral blocks} $B$ denote a class of semantic operation rather than a concrete action. For example, communicating with a user is one behavior, while actions such as \texttt{WELCOME}, \texttt{EXPLAIN}, or \texttt{WARN} are values of that behavior. The same behavioral block can be present multiple times in a workflow, as different agents may perform it differently. For example, there can be several Analyzer agents who take the same signal but apply different methods to process it (i.e., ensemble learning). Typical behaviors in our workflows include observing, assessing, planning, gating for safety, executing actions, communicating with users, and enforcing policies. This abstraction \textit{avoids proliferating narrowly defined blocks that differ only in parameter values. Instead, workflows are composed from a small set of reusable semantic operations}. 

\textbf{Execution edges} $E$ define the operational dependencies between blocks during a single workflow step. An execution edge $(b_i, b_j)$ indicates that the output of block $b_i$ provides input to block $b_j$. These edges form the execution graph of the workflow. The execution graph is constrained to be acyclic, e.g. a moderation workflow may observe a post, assess its content, produce a decision plan, pass the plan through a safety gate, and finally execute the resulting actions. These dependencies represent the operational pipeline that a workflow engine executes during a given cycle. Workflows also contain \textit{state coupling relationships}. A block may modify the persistent state of an agent, thus affecting \textit{future} behavior rather than the current execution step. For instance, a policy enforcement block may update an account's status to \texttt{RESTRICTED} or \texttt{BANNED}. This write does not immediately trigger further execution in the same cycle; instead, it constrains which behaviors will be available to the agent in subsequent cycles. Distinguishing execution flow from state coupling preserves feedback semantics without creating cyclic control graphs. The workflow is acyclic as an execution process, while feedback occurs through state updates that influence future interactions.

A \textbf{guard} specifies when a block may execute or a transition may occur. For example, on a post-creation block, $[\text{status} \neq \text{BANNED}]$ permits execution only when the account is not banned. Guards allow workflows to encode context-sensitive behavior in which actions are enabled or disabled depending on the state of agents or the environment. This mechanism mirrors how actions in ABMs are conditional on internal states and contextual conditions.

On the basis of this formalism, we developed a set of 12 rules (Table \ref{tab:workflow_rules}) to detect common design flaws in agentic AI workflows. The rules are designed to strike a balance: overly generic rules may detect issues without providing insight into what is wrong or how to fix it, while extremely case-specific rules would require enumerating every possible problem that a workflow designer could make. Instead, our rules capture recurring \textit{categories of issues} that arise when composing agentic workflows. We do not claim that this set constitutes a definitive or complete framework for verifying all agentic AI systems\footnote{Our focus on design-time structural composition means that we do not inspect the content of prompts or tool invocations, so several documented categories of failures are outside of our scope. For instance a workflow can have valid message types, gates, and appeal paths while an LLM still follows malicious instructions embedded in user content or retrieved documents as studied by \shortcite{debenedetti2024agentdojo,zhan2024injecagent}. The verifier can require a policy gate before enforcement but it does not decide whether an agent should refuse a malicious multi-step request. AgentHarm shows that harmful agentic tasks such as cybercrime and harassment require separate misuse evaluation \shortcite{andriushchenko2025agentharm}. Finally our rules check whether a tool-using block is connected coherently but they do not inherently verify whether the tool call itself is safe, reversible, sandboxed or authorized at runtime. Prior works like ToolEmu document such failures for tool-using agents \shortcite{ruan2024identifying}.}. As discussed in the limitations section, empirically collecting a larger corpus of real-world agentic workflows and systematically analyzing their design failures would be necessary to expand and refine the rule set. Note that rules may interact, as violations can co-occur or compound even though each rule is checked independently. This issue has already been studied in rule-based verification, where conflict and dependency analysis studies interactions among rules \shortcite{lambers2019granularity}. \textcolor{black}{Table~\ref{tab:rule_interactions} exemplifies several interactions among our rules.}

\begin{table*}[htb]
\centering
\caption{We have 12 verification rules covering four categories of issues in Agentic AI workflows. Rules and rationales are detailed in our online supplementary material (\url{https://doi.org/10.5281/zenodo.19502076})}
\label{tab:workflow_rules}
\small
\begin{tabularx}{\textwidth}{p{1.3cm} c X}
\toprule
\textbf{Category} & \textbf{\#} & \textbf{What the rule checks and why it is needed} \\
\midrule

\multirow{3}{*}{\shortstack{Evidence \&\\quality}}
& 1 & State-changing actions (e.g., updating trust, restricting users, reverting edits) must trace back to an upstream block that produces interpreted evidence, otherwise changes lack observable causes \\

& 2 & Raw external inputs (e.g., posts, revisions) must be evaluated (through a specialized agent) before triggering state changes, otherwise ambiguous events are treated as final evidence \\

& 3 & Multiple sources of evidence (e.g., automated moderation and user reports) must be consolidated into one observation before updating state, to avoid double-counting or contradictory updates \\

\midrule

\multirow{4}{*}{\shortstack{Decision \&\\governance}}
& 4 & Strong actions (e.g., bans, reverts, restrictions, page locks) must pass through one clear point for decision or policy gate rather than being triggered directly from observations \\

& 5 & Blocks with prerequisites (e.g., account state, case history) must have those inputs upstream in the workflow, or else planners/ enforcers make decisions without the information they require \\

& 6 & Blocks acting under a single authority (such as executors or enforcers) must have exactly one control source. Otherwise multiple independent paths attempt to trigger the same action \\

& 7 & Blocks depending on prior history (e.g., escalation planners, loop-safety checks) must have access to the relevant feedback or state, otherwise history-dependent decision cannot be made \\

\midrule

\multirow{3}{*}{\shortstack{Message \&\\consistency}}
& 8 & Every connection must send a message type that the destination block explicitly accepts, so workflows respect the semantics of observations, decisions, plans, and other message types \\

& 9 & The identity of incoming messages must match the entity targeted by the action (e.g., actions affecting a reported author must use the author identity rather than the reporter identity) \\

& 10 & Blocks must receive all contextual inputs needed for their semantic role (e.g., editor history, case records, page status) or they cannot meaningfully perform the decision they claim to implement \\

\midrule

\multirow{2}{*}{\shortstack{Appeal \&\\safeguards}}
& 11 & Appeal/review endpoints must be reachable from external appeal inputs even when enforcement actions restrict normal interaction paths, otherwise there is no procedural recourse \\

& 12 & Nodes that detect the need for escalation or review must connect to a downstream review or mediation endpoint, otherwise we provide no resolution mechanism \\

\bottomrule
\end{tabularx}
\end{table*}

\begin{table*}[htb]
\centering
\caption{\textcolor{black}{Representative interactions among verification rules. These are not additional rules, they show how independently-detected violations can compound in a workflow.}}
\label{tab:rule_interactions}
\small
\begin{tabularx}{\textwidth}{p{2cm} p{4.3cm} X}
\toprule
\textbf{Co-triggered rules} & \textbf{Interaction} & \textbf{Practical implication} \\
\midrule
1 + 4 & Ungrounded state change and missing decision gate & State updates or enforcement actions may occur without both interpreted evidence and policy mediation. \\
3 + 4 & Unresolved evidence streams and missing decision gate & Conflicting observations can drive a strong action before being canonicalized and authorized. \\
6 + 7 & Competing control sources and missing feedback/history & A history-dependent block can be triggered by multiple authorities while lacking the state needed to contextualize the decision. \\
8 + 9 & Message-contract mismatch and identity mismatch & A workflow may route an incompatible message and bind the resulting action to the wrong entity. \\
11 + 12 & Unreachable appeal endpoint and missing escalation path & A workflow may detect the need for review while still failing to provide reachable procedural recourse. \\
\bottomrule
\end{tabularx}
\end{table*}

\subsection{Evaluating Rule Detection on Three Case Studies}

To evaluate the proposed rules, we defined a \textit{reference workflow} for each use case. Each reference workflow (Figure~\ref{fig:workflows-overview}) represents a well-designed decision pipeline in which no rule should be triggered. To test whether rules can detect design flaws, we created two sets of deviations from the reference workflows. In the first set (called `errors'), we created variants of each workflow by intentionally introducing one, two, or three errors. By precisely designing this set, we have ground truth information on how many violations should be detected and where they are caused. We ensured that this set is balanced to avoid reporting biased performance estimates, thus combining 24 correct workflows with 24 incorrect ones (12 single errors, 7 double errors, 5 triple errors). Each of the three case studies contributes in the same proportion to this set. 

The second set evaluates the robustness of the verifier when workflows are represented differently but preserve the same underlying logic. For examples, designers can have intermediate agents, or agents performing the same role may be split or merged. These transformations create workflows that are structurally different but contain the core logic operations of the originals. We applied such transformations to create 168 workflows. To support replicability and transparency, our two datasets along with the implemented verification rules are permanently accessible on a third-party repository at \url{https://doi.org/10.5281/zenodo.19504756}.



\section{RESULTS}
\label{sec:results}

\subsection{Software Prototype}

\begin{figure}[htb]
\centering
\includegraphics[width=1\textwidth]{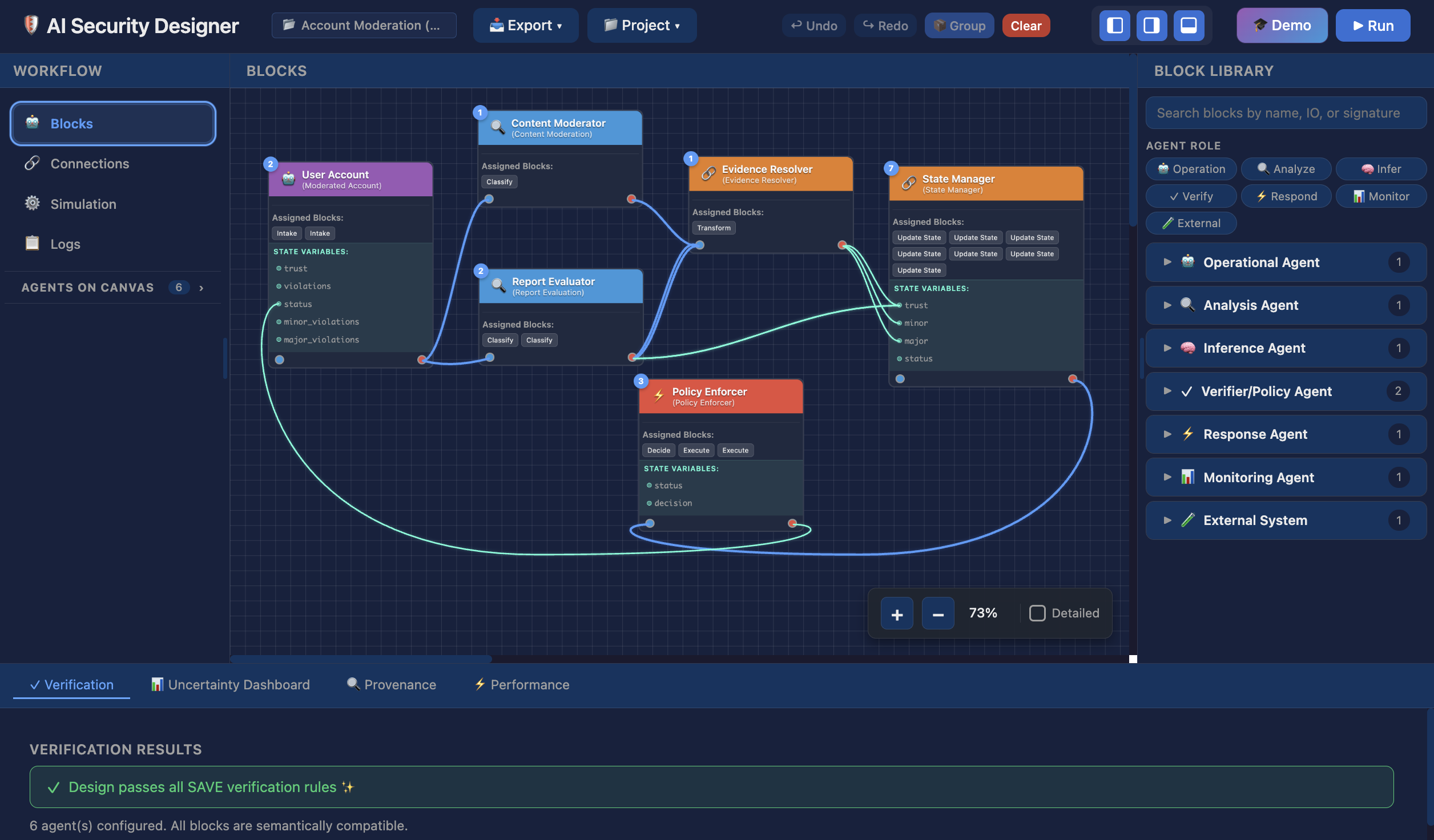}
\caption{Screenshot of our software. A demonstration is available at \url{http://youtu.be/MzQDpK8CXWE}}
\label{fig:software_overview}
\end{figure}

To demonstrate the feasibility of our approach, we developed a browser-based software that embodies our verification framework. Our software prototype allows users to create agentic AI workflows and verify them during the design phase. The interface consists of four panels (Figure~\ref{fig:software_overview}), from left to right: project management (e.g., working on blocks or running simulations), canvas to compose workflows, reusable behavioral building blocks organized by agent role, and a bottom panel to analyze workflows' correctness. The canvas distinguishes between the two types of execution edges (Section~\ref{sec:representation}): control flow from one block to another (shown in thick blue lines) and state changes from one block to specific state variables (shown in thinner green lines). The interface provides several functionalities to support productivity, e.g. several agents can be grouped (to deal with large workflows and/or situations in which some parts of the workflow are stable while others are ideated through design), the library of building blocks is searchable and provides filters, and details-on-demand can show detailed properties (either by double-click or enabling `Detailed' in the bottom right of the canvas). When a workflow triggers one of our rules during verification, the bottom panel explains what should be fixed, proposes an automatic fix, and shows the user which part of the workflow is affected. To support integration with existing pipelines, a command-line interface and a programmatic API allow external tools to upload workflow, trigger their verification, and retrieve fixes.

\subsection{Evaluation}
We first evaluated our design time verifier on the `error' set of 48 workflows. As the verifier is deterministic and the faulty workflows were generated by systematically introducing known structural violations, the experiment evaluates implementation correctness for the specified rule set\footnote{We check whether the software emits the expected rule identifiers and violation counts for controlled workflows. The experiments do not evaluate whether our 12 rules characterize all real-world design flaws. Practical workflows may contain safety-relevant flaws beyond our rules (Table~\ref{tab:workflow_rules}). Creating a broader set of cases is an important goal for benchmarking.}. On this set, the verifier correctly identified all expected violations, yielding precision, recall, and F1 equal to 1.000 and zero count error (MAE = 0, RMSE = 0). Performance was consistent across the four benchmark categories (correct, single-error, double-error, and triple-error) and across the three use cases. We then evaluated the verifier on the 168 workflow variants that preserve the core logic but use different graph representations. Again, the verifier detected the expected violations in all cases. This result indicates that rule detection does not depend on a specific encoding of the workflow graph and remains stable under equivalent structural transformations.

\section{DISCUSSION}
\label{sec:discussion}

This paper addresses the growing need for design-time verification of agentic AI workflows. Our work contributes a set of 12 cross-cutting verification rules that capture recurring design issues, together with a formal representation and a software prototype capable of detecting these issues during system design. Through three case studies and two sets of workflows, we showed that these rules can reliably identify violations even when workflow logic is represented through different graph structures (e.g., introducing intermediate agents or splitting responsibilities across multiple nodes). This result suggests that design-time verification can operate at the level of workflow composition, even when design issues may be masked through superficial structural variations such as agent splitting or redundant agents.

\textcolor{black}{The 12 rules should thus be read as a partial design-time rule set, not as a complete safety specification for real agentic workflows. They cover flaws expressible in our representation as incompatible or missing relationships among agents, messages, state variables, guards, behavioral blocks, and execution edges. They do not cover safety failures that depend on unmodeled prompt content, tool sandboxing, model behavior, adversarial data, organization-specific policy choices, or domain norms. Following the distinction made in compositional verification between local component contracts and global system guarantees \cite{chilton2014compositional}, our claim is that the current rules make a defined class of composition errors checkable while completeness over real workflows remains an empirical research problem requiring larger failure corpora and community refinement of the rule library.}

Our work faces two main limitations. First, there is a lack of benchmarks documenting design flaws in real-world agentic AI workflows. We partially addresses this gap by constructing and releasing a curated dataset of workflow variants with labeled violations, which can serve as an initial benchmark for evaluating design-time verification methods. However, this dataset remains necessarily limited in scope because it was derived from three case studies. Second, we rely on a library of reusable building blocks. \citeN{kasputis2000composable} noted that such  libraries must be sufficiently rich to support multiple applications, which is a major undertaking. As envisioned in prior M\&S work on composable simulations \cite{balci2011achieving}, developing repositories of reusable components is a community-level effort that extends beyond the scope of any single study. In the context of agentic AI, new coordination patterns, safeguards, and mechanisms will continue to emerge as systems evolve. Community-driven initiatives such as the {\ttfamily AGENTBLOCKS} platform illustrate \textcolor{black}{the governance model needed for this ecosystem: reusable blocks can be maintained in an open, version-controlled registry (e.g., a GitHub repository linked to a web catalog) where submissions include typed inputs and outputs, assumptions, semantic constraints, and compatibility tests, and are reviewed by maintainers \shortcite{filatova2025agentblocks}.} \textcolor{black}{A practical registry could apply semantic versioning to block interfaces and constraints, and expose certification tiers such as \textit{experimental}, \textit{reviewed}, and \textit{benchmark-validated}. We thus do not assume that a single study can guarantee long-term maintenance. Rather, we defined verification constraints that can be reused with community-maintained blocks as such registries evolve.}

\section*{ACKNOWLEDGMENTS}
This work was supported by the
Coastal Virginia Center for Cyber Innovation (COVA CCI) Commonwealth Cyber Initiative
(CCI), an investment in the advancement of cyber research \& development, innovation and
workforce development. For more information about CCI, visit \href{www.covacci.org}{covacci.org} and
\href{www.cyberinitiative.org}{cyberinitiative.org}.

\footnotesize

\bibliographystyle{wsc}

\bibliography{demobib}

\section*{AUTHOR BIOGRAPHIES}

\noindent {\bf \MakeUppercase{Noé Flandre}} holds a Master’s degree in Computer Science and is an engineering graduate from IMT Mines Alès (France). His research interests lie at the intersection of Modeling \& Simulation and Artificial Intelligence. His work has appeared in the Journal of Simulation and at the International Conference on Conceptual Modeling. His e-mail address is \email{noeflandre@gmail.com}.\\

\noindent {\bf \MakeUppercase{Alexander C. Nwala}} is an Assistant Professor of Data Science at the College of William \& Mary. His research focuses on the analysis of social and information systems, including the detection and characterization of social bots and the modeling of online account behavior. His e-mail address is \email{acnwala@wm.edu}.\\

\noindent {\bf \MakeUppercase{Philippe J. Giabbanelli}} is a Research Professor at Old Dominion University in Norfolk, Virginia. He has authored over 180 articles (mostly on Modeling \& Simulation and AI applied to human behavior), of which 15 have appeared at the Winter Simulation Conference, where he co-leads the professional development track. His e-mail address is \email{pgiabban@odu.edu}.\\

\end{document}